# Mamba-UIE: Enhancing Underwater Images with Physical Model Constraint

Song Zhang [a, b, c, d], Yuqing Duan [a, b, c, d],

Daoliang Li [a, b, c, d], Ran Zhao [a, b, c, d *]

15620503230@163.com; Duanyuqing0450@163.com;
dliangl@cau.edu.cn; ran.zhao@cau.edu.cn

[a] National Innovation Center for Digital Fishery, China Agricultural University, Beijing 10083, China

[b] Key Laboratory of Smart Farming Technologies for Aquatic Animal and Livestock, Ministry of Agriculture and Rural Affairs, China Agricultural University, Beijing 10083, China

[c] Beijing Engineering and Technology Research Center for Internet of Things in Agriculture, China Agricultural University, Beijing 10083, China

[d] College of Information and Electrical Engineering, China Agricultural University, China Agricultural University, Beijing 10083, China

* Corresponding author at: P. O. Box 121, China Agricultural University, 17 Tsinghua East Road, Beijing 100083, China. E-mail address: ran.zhao@cau.edu.cn (R. Zhao)

**Abstract:** In underwater image enhancement (UIE), convolutional neural networks (CNN) have inherent limitations in modeling long-range dependencies and are less effective in recovering global features. While Transformers excel at modeling long-range dependencies, their quadratic computational complexity with increasing image resolution presents significant efficiency challenges. Additionally, most supervised learning methods lack effective physical model constraint, which can lead to insufficient realism and overfitting in generated images. To address these issues, we propose a physical model constraint-based underwater image enhancement framework, Mamba-UIE. Specifically, we decompose the input image into four components: underwater scene radiance, direct transmission map, backscatter transmission map, and global background light. These components are reassembled according to the revised underwater image formation model, and the reconstruction consistency constraint is applied between the reconstructed image and the original image, thereby achieving effective physical constraint on the underwater image enhancement process. To tackle the quadratic computational complexity of Transformers when handling long sequences, we introduce the Mamba-UIE network based on linear complexity state space models. By incorporating the Mamba in Convolution block, long-range dependencies are modeled at both the channel and spatial levels, while the CNN backbone is retained to recover local features and details. Extensive experiments on three public datasets demonstrate that our proposed Mamba-UIE outperforms existing state-of-the-art methods, achieving a PSNR of 27.13 and an SSIM of 0.93 on the UIEB dataset. Our method is available at https://github.com/zhangsong1213/Mamba-UIE.

**Keywords**: Underwater Image Enhancement, Revised Underwater Image Formation Model, Mamba, Reconstruction Consistency

## 1. Introduction

With the continuous exploration of marine environments, underwater imaging technology is playing an increasingly important role in visual tasks such as autonomous navigation [1], underwater target detection [2], and environmental monitoring [3]. However, due to the unique underwater environment, light inevitably undergoes attenuation as it propagates underwater, leading to the degradation and distortion of underwater images [4]. This poses significant challenges for vision-based tasks. Improving the recognizability and contrast of targets in underwater images will undoubtedly reduce the burden on downstream visual tasks. Therefore, utilizing underwater image enhancement (UIE) to restore and improve the visual quality of underwater images is crucial for enhancing the performance of downstream tasks.

UIE is a classic yet challenging task in the field of computer vision. The unavailability of real ground truth for underwater images makes UIE an ill-posed problem. Traditional direct methods [4-7] operate directly on pixels using algorithms to correct colors and enhance contrast. Priors-based methods model the underwater imaging process based on prior knowledge, estimating model parameters to reverse the degradation process and obtain high-quality underwater images [8-10]. Both direct and priors-based methods suffer from low robustness and often exhibit over-enhancement in complex underwater scenarios.

In recent years, due to the rapid development of neural networks and their powerful representational capabilities, researchers have introduced convolutional neural network (CNN) to achieve nonlinear mappings from low-quality underwater images to high-quality ones [11-13]. CNN-based UIE has demonstrated impressive performance, as it can automatically learn image features and accomplish end-to-end learning. However, CNNs are limited by their restricted receptive field size, making them ineffective at capturing long-range dependencies. To address this, researchers have proposed Transformer based on self-attention mechanism to effectively extract global features [14-16]. Nonetheless, the quadratic computational complexity exhibited by Transformer when processing long sequences poses challenges for high-resolution image processing. Additionally, most supervised learning methods typically rely on the loss function between outputs and labels to constrain the model, lacking effective physical model constraints. This often results in generated images lacking realism and leads to over-enhancement.

Recently, state space models (SSM) [17-19], such as Mamba [20], have been proposed to address the quadratic computational complexity of Transformer. These models have gained widespread attention and application in visual tasks such as semantic segmentation [21] and object detection [22]. This provides new ideas and methods for solving underwater image enhancement problems. To address the aforementioned issues, we propose a physically model-constrained underwater image enhancement framework, Mamba-UIE. Specifically, we decompose the input image into four components: scene radiance, direct transmission map, backscatter transmission map, global background light. These components are then reassembled according to a revised underwater image formation model, with a reconstruction consistency constraint applied between the reconstructed image and the original one. This achieves effective physical constraints on the underwater image enhancement process. Additionally, to address the quadratic computational complexity issue of Transformer when processing long sequences, we propose the Mamba-UIE network based on linear complexity SSM (note that both our proposed framework and network are named Mamba-UIE). This is a parallel processing network incorporating both CNN and Mamba, using the Mamba in Convolution (MIC) block to model long-range dependencies at both channel and spatial levels, while retaining the CNN backbone to recover local features and details. Fig. 1 shows several enhanced images generated by Mamba-UIE. It is evident that Mamba-UIE achieves visually pleasing results in terms of color, contrast, and naturalness.

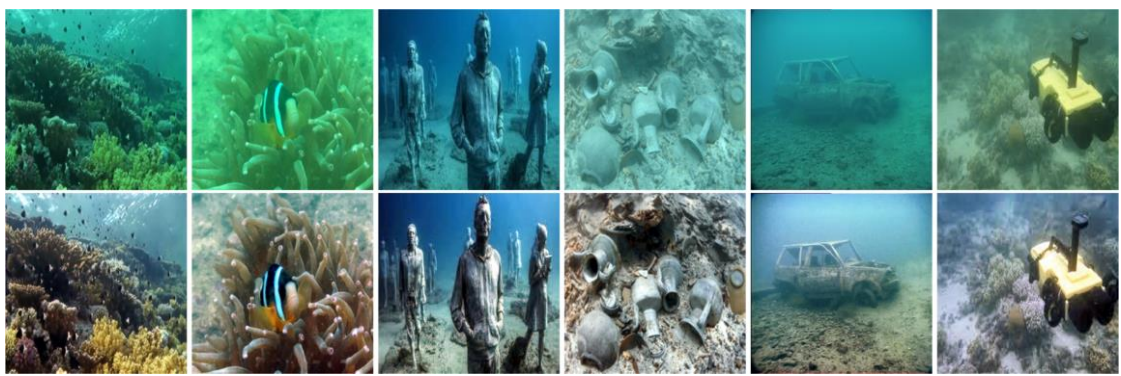

**Fig. 1**. Results of the proposed Mamba-UIE on common underwater visual scenes. (Top) Raw underwater images of coral reefs, fish, statues, jars, sunken cars, AUV. (Bottom) Corresponding enhanced images generated by Mamba-UIE.

We conducted extensive qualitative and quantitative experiments and analyses on three public datasets, demonstrating the effectiveness of the proposed Mamba-UIE. The main contributions of this paper are as follows:

1. We propose a physical model-constrained underwater image enhancement framework. By incorporating a physical formation model, we consider physical laws and real-world characteristics in the process of image formation, improving the realism and robustness of the generated images. This framework is simple, effective, and easily extendable to other formation models.
2. We introduce a parallel processing network combining CNN and Mamba. By incorporating the MIC block, we achieve modeling of long-range dependencies at both channel and spatial levels, while retaining the CNN backbone for recovering local features and details.
3. Extensive experiments demonstrate that our proposed Mamba-UIE outperforms the current state-of-the-art models. This provides a powerful and promising solution for UIE.

## 2. Related works

### 2.1 Transformer-based UIE

In recent years, CNN-based UIE [12, 13, 23, 24], has developed rapidly. Compared to traditional direct methods [7, 25-27] and priors-based methods [10, 28], CNN can automatically learn effective features from underwater images through end-to-end learning. However, the local receptive field mechanism of CNNs restricts them to focus only on local information in the image, making it difficult to capture global information. Solving this problem by increasing the receptive field leads to a significant increase in computational complexity. These limitations hinder the further development of CNNs in the field of image enhancement.

To achieve modeling of global information, Transformer-based UIE has gained attention in recent years [29-32]. Compared to CNNs, the self-attention mechanism in Transformers can capture long-range dependencies in the image, overcoming the limitations of the local receptive field inherent in traditional CNNs. Ren, et al [29]. incorporated the Swin Transformer [14] into U-Net to enhance the ability to capture global dependencies. Considering the insufficiency of Transformers in capturing local attention, they integrated CNN with core attention mechanism to capture more local attention. Similarly, Wang, et al [31]. proposed a feature fusion Transformer for global information fusion and reconstruction, combining CNN to achieve both global and local modeling of underwater images. Shen, et al [30]. introduced a dual-attention mechanism by combining channel self-attention Transformer and pixel self-attention Transformer. Taking into account the prior knowledge that underwater images degrade unevenly in different spatial regions, Peng, et al. [32] designed a novel spatial-wise global feature modeling Transformer based on spatial self-attention mechanisms to enhance the network's focus on more severely attenuated regions. They also proposed a U-shape Transformer. Despite the ability of visual Transformers to establish strong global dependencies in UIE, the quadratic computational complexity brought by self-attention poses new challenges as image resolution increases.

**2.2 State space model**

State Space Models (SSM) [17-19], originating from classical control theory, have recently been introduced into deep learning as a competitive backbone for state space transformations. The promising property of linearly scaling with sequence length in long-range dependency modeling has attracted great interest from searchers. The Structured State Space Model (S4) [18] is a pioneering deep state space model in modeling long-range dependencies. Subsequently, the S5 layer [19] was proposed based on S4, introducing multi-input, multi-output (MIMO) SSM and efficient parallel scanning. Mamba [20] is a data-dependent SSM featuring a selective mechanism and efficient hardware design, which scales linearly with input length and offers lower computational complexity compared to Transformer.

Due to Mamba's outstanding capability in modeling long-range dependencies, it has gradually been applied in the visual tasks, such as medical image segmentation [33, 34], image restoration [35], and image classification [36, 37]. Recently, Mamba has also been applied in the UIE. Guan, et al. [38] proposed an SSM-based UIE method, WaterMamba, to address the quadratic computational complexity problem of Transformers. They embedded spatial-channel coordinate omnidirectional selective scan into U-Net to reconstruct image details and colors. Chen, et al. [39] designed a dynamic interaction-visual SSM to model global dependencies. Lin, et al. [40] proposed PixMamba, which utilizes a dual-layer network to extract both global and local feature information. These methods use Mamba for UIE, establishing global dependencies, which may lead to neglecting the modeling of local relationships. Therefore, this paper proposes combining Mamba with CNN to achieve both global and local dependency modeling.

## 2.3 Physical model constraint-based UIE

In the aforementioned deep learning methods, an end-to-end training approach is employed, directly learning the mapping relationship from inputs to labels. The effectiveness of this approach largely depends on how well the loss function constrains the model.

In unsupervised learning, physical prior models are fully utilized. Yan, et al. [41] proposed a model-driven cycle-consistency GAN that alleviates underwater image degradation by estimating environmental variables. Yan, et al. [42] employed neural networks to model underwater scene, estimating scene depth, scatter, attenuation and veiling light to compute underwater scene radiance. Huang, et al. [43] designed a color mapping network and a contrast mapping network, using the gray-world assumption and dark channel prior (DCP) as constraints. From the perspective of layer disentanglement, Chai, et al. [44] decomposed underwater images into different components based on physical models and applied constraints to the reconstructed images. Similarly, Khan, et al. [45] decomposed underwater images into reflection and illumination components. Based on the Koschmieder model, Kar et al. [46] found that appropriate degradation of the input image is equivalent to a controlled perturbation describing image formation. They proposed a zero-shot unsupervised training framework. Similarly, Fu, et al. [47] utilized a homology consistency between re-degraded image and input image to apply appropriate constraints to the generated clean images. In supervised learning, high-quality labels are available. By combining physical models, neural network models can be further effectively constrained, generating more natural and high-quality enhanced underwater images.

## 3. Proposed method

## 3.1 Preliminaries

### 3.1.1 Revised underwater image formation model

Our method is based on the revised underwater image formation model [48], which describes the degradation of underwater images caused by light scattering and absorption. According to the Beer-Lambert law [49], the propagation of light is related to the attenuation factor $e^{-\beta d(x)}$, where $d$ represents the distance to the light source. $\beta(>0)$ is the channel extinction coefficient, which increases with the density of scattering-induced particles. According to the Beer-Lambert law, the light scattering effect at a distance $d$ from the light source in a water medium can be expressed as:

$$B(x) = (1 - e^{-\beta d(x)})A \tag{1}$$

where $A$ represents the global background light, and $B(x)$ is known as backscattering, which results from the reflection of light by suspended particles causing image degradation. The scene radiance $J$ is also affected by the attenuation factor $e^{-\beta d(x)}$. Thus, we have:

$$D(x) = J(x)e^{-\beta d(x)} \qquad (2)$$

where $D(x)$ contains the direct signal of the scene. The underwater image formation model is typically represented as the sum of $B(x)$ and $D(x)$. However, this neglects the impact of light absorption on the attenuation coefficient.

Akkaynak, et al. [48] observed that in underwater scenes, the direct signal and backscattering are affected by two different attenuation coefficients, namely, the RGB attenuation and backscattering coefficients. To achieve a more accurate model, they revised the underwater image formation model as follows:

$$I(x) = J(x)e^{-\beta^D d(x)} + (1 - e^{-\beta^B d(x)})A \qquad (3)$$

where $I(x)$ represents the underwater image, $J(x)$ is the scene radiance, $x$ is the pixel in the image, $\beta^D$ and $\beta^B$ are the RGB attenuation coefficient and backscattering coefficient, respectively. Due to the wavelength selection characteristics, each color channel has different RGB attenuation and backscattering coefficients. For simplicity, we simplify Eq. (3) as follows:

$$I(x) = J(x)T_D(x) + (1 - T_B(x))A \qquad (4)$$

where $T_D(x) = e^{-\beta^D d(x)}$ represents the direct transmission map, and $T_B(x) = e^{-\beta^B d(x)}$ represents the backscattering transmission map.

### 3.1.2 State space model

The recent advancements in structured State Space Sequence Models (S4) are largely inspired by continuous Linear Time-Invariant (LTI) systems. These models map a one-dimensional function or sequence $x(t) \in \mathbb{R}$ to $y(t) \in \mathbb{R}$ through an implicit latent state $h(t) \in \mathbb{R}^N$. Formally, this can be represented using linear ordinary differential equation (ODE) as follows:

$$\begin{aligned} h'(t) &= \mathbf{A}h(t) + \mathbf{B}x(t), \\ y(t) &= \mathbf{C}h(t) + \mathbf{D}x(t) \end{aligned} \qquad (5)$$

where $\mathbf{A} \in \mathbb{R}^{N \times N}$, $\mathbf{B} \in \mathbb{R}^{N \times 1}$, $\mathbf{C} \in \mathbb{R}^{1 \times N}$, $\mathbf{D} \in \mathbb{R}$, with $N$ being the state size.

A discretization process is typically used to integrate Eq. (5) into practical deep learning tasks. Specifically, a time scale parameter $\Delta$ is used to transform the continuous parameters $\mathbf{A}$ and $\mathbf{B}$ into discrete parameters $\overline{\mathbf{A}}$ and $\overline{\mathbf{B}}$. A common

discretization method is the zero-order hold (ZOH) rule, which can be defined as follows:

$$\overline{A} = \exp(\Delta A),$$
$$\overline{B} = (\Delta A)^{-1}(\exp(A) - I) \cdot \Delta B \qquad (6)$$

where $I$ represents the identity matrix. After discretization, the discretized Eq. (5) with step size $\Delta$ can be rewritten in the form of a recurrent neural network (RNN) as follows:

$$h_k = \overline{A}h_{k-1} + \overline{B}x_k,$$
$$y_k = Ch_k + Dx_k \qquad (7)$$

Furthermore, Eq. (7) can be mathematically equivalently expressed in the form of a CNN as follows:

$$\overline{K} \triangleq (C\overline{B}, C\overline{AB}, ..., C\overline{A}^{L-1}\overline{B}),$$
$$y = x * \overline{K} \qquad (8)$$

where $L$ is the length of the input sequence, $*$ denotes the convolution operation, and $\overline{K} \in \mathbb{R}^L$ is the structured convolution kernel.

The state-of-the-art state space model Mamba [20] further improves $\overline{B}$, $C$ and $\Delta$, to make them input-dependent, thereby allowing dynamic feature representation. Specifically, Mamba has the same recursive form as Eq. (7), enabling the model to remember extremely long sequences. Additionally, the parallel scan algorithm [20] provides Mamba with the same parallel processing advantages as Eq. (8), facilitating efficient inference.

**3.2 Overall architecture**

Our proposed Mamba-UIE utilizes the revised underwater image formation model to effectively constrain the network. In the formation model, considering the impact of light absorption on the attenuation coefficient, the direct transmission map and the backscattering transmission map are distinguished. The backscattering transmission map is one of the main causes of underwater image blurriness. Compared to the Koschmieder light scattering model [50], the revised underwater image formation model performs better (as detailed in the Section 4.6 Ablation Study). As shown in Fig. 2, we decompose the underwater image $I(x)$ into four latent components: global background light $A$, direct transmission map $T_D(x)$, backscattering transmission map $T_B(x)$, and scene radiance $J(x)$ (i.e., the clean underwater image to be recovered). Effective regularization constraints are applied between the scene radiance and the ground truth. Additionally, in accordance with the revised formation model, the components are reorganized to obtain the reconstructed image $I'(x)$:

$$I'(x) = J(x)T_D(x) + (1 - T_B(x))A \qquad (9)$$

If these estimated components are accurate, the reconstructed image $I'(x)$ will be identical to the original image $I(x)$. Similarly, applying a reconstruction consistency constraint between the original and reconstructed images will improve the accuracy of the estimated components. The specific reconstruction consistency constraint is detailed in Section 3.6 Loss Function.

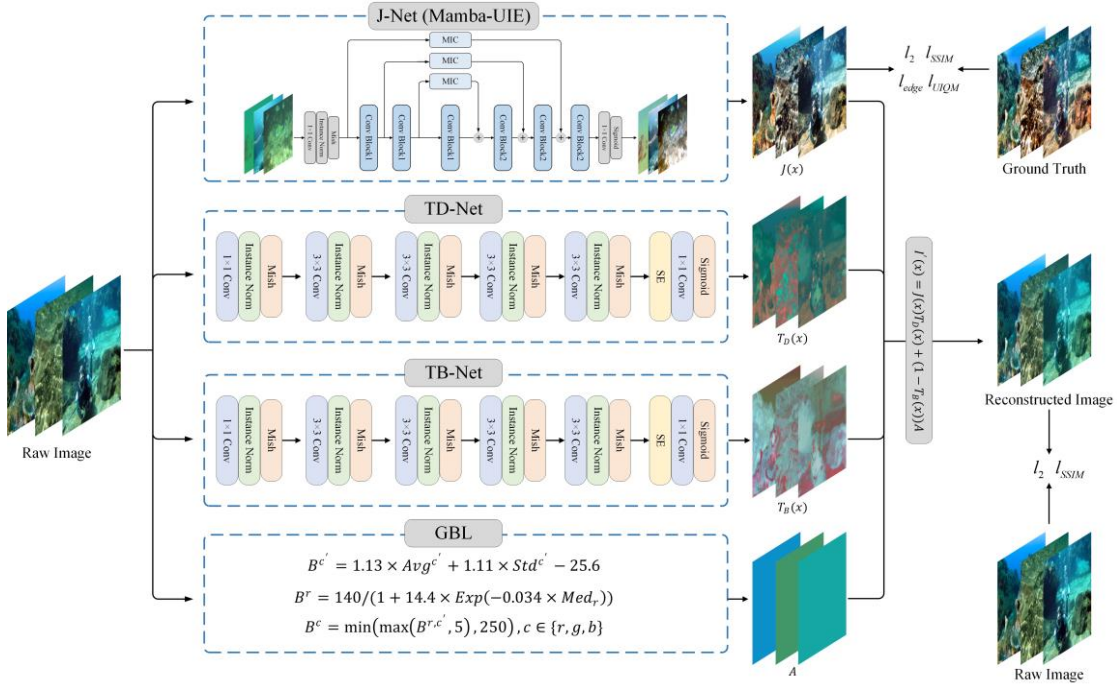

**Fig. 2**. Our proposed Mamba-UIE framework, which consists of a scene radiance estimation network (J-Net), a direct transmission map estimation network (TD-Net), a backscatter transmission map estimation network (TB-Net), and a global background light estimation module (GBL). These modules are used to estimate the corresponding components.

### 3.3 Mamba-UIE

To achieve better global dependency relationships, we designed a Mamba-based scene radiance estimation network, Mamba-UIE. In previous underwater image enhancement efforts, Transformer was often used to extract global features. While effective, Transformer has the drawback of high computational complexity. Mamba effectively addresses this issue. Considering Mamba's advantages in global feature extraction but its shortcomings in detail recovery, we integrated CNN, whose inductive biases favor image detail recovery. Therefore, we introduced Mamba as a parallel module to enhance the network model's capability in modeling long-range dependencies. In our network design, we adopted a hybrid architecture combining Mamba and CNN. This approach ensures effective extraction of global features while also guaranteeing efficient recovery of image details. The proposed Mamba-UIE network is illustrated in Fig. 3.

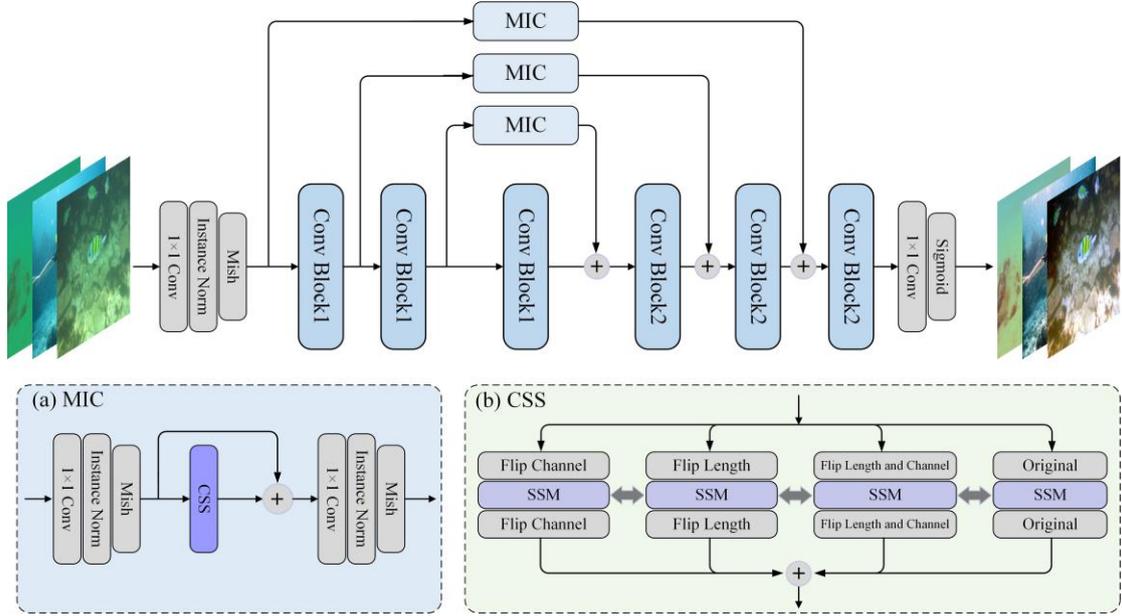

**Fig. 3**. Our proposed Mamba-UIE network includes the (a) Mamba in Convolution (MIC) module and the (b) Channel-Spatial Siamese Learning (CSS) module. Conv represents convolution.

In Mamba-UIE, the backbone uses a series of convolutions and Conv Blocks to extract local features from the input. The first convolution is used to quickly increase the number of channels. Conv Block1 comprises a 3×3 Conv, an Instance Normalization (IN) layer, and a Mish activation function. This section is responsible for extracting both shallow and deep features of the image while transmitting these features to the MIC module. Conv Block2 includes a Squeeze-and-Excitation (SE) layer, a 3×3 conv, an IN layer, and a Mish activation function. This part primarily selects features from the local features provided by the CNN and the global features from the MIC module, along the channel dimension. The final convolution converts all feature information to the RGB channels while normalizing the pixel values between 0 and 1. Finally, the pixel values are mapped to the 0-255 range to obtain the final enhanced image. Additionally, our proposed Mamba-UIE is a non-degrading network structure, meaning it does not perform downsampling operations on the image. This effectively prevents the loss of details during the image enhancement process.

**MIC:** Inspired by the Network in Network (NIN) [51], MIC inputs features from the convolution layer into Mamba to achieve better nonlinear long-range modeling, while also retaining the feature inputs from the convolution. As shown in Fig. 3(a), MIC

can be simply described as:

$$F_{out} = Conv.O(CSS(Conv.I(F_{in})) + Conv.I(F_{in})) \qquad (10)$$

where, $F_{in}$ and $F_{out}$ represent the input and output features, respectively. $Conv.I$ and $Conv.O$ represent the input and output Conv Blocks, each consisting of a 1×1 conv, an IN layer, and a Mish activation function.

**CSS:** The core advantage of the NIN method lies in using Multilayer Perceptions (MLP) to encapsulate the nonlinear capabilities of features; however, MLPs struggle to handle the flattened features in $C \times H \times W$ dimensions. On the other hand, SSM excels at modeling the length relationships within sequences. As shown in Fig. 3(b), the CSS module reshapes the input features along the channel and spatial dimensions, utilizing SSM to achieve better long-range dependencies and improved interaction between channel and spatial features.

### 3.4 Transmission map estimation modules

The propagation of light underwater is influenced by water quality, depth, and suspended particles, with different wavelengths experiencing varying degrees of attenuation. The direct transmission map reflects the extent of scattering that occurs during light transmission, while the backscattering transmission map indicates the proportion of scattered light returning to the camera. Together, these two maps determine the final image quality. For estimating these transmission maps, we use identical networks, named TD-Net and TB-Net, respectively, as shown in Fig.2. TD-Net and TB-Net each contain six convolutional layers. The first layer uses a 1×1 convolution kernel to quickly increase the number of channels. The last convolutional layer incorporates a SE layer for feature selection along the channel dimension. The output is achieved through a 1×1 convolution followed by a Sigmoid function, with the final values mapped back to the 0-255 range. Notably, TD-Net and TB-Net do not have explicit constraints; they are optimized through the reconstruction loss based on the top layer of the framework.

### 3.5 Global Background light estimation module

The global background light exists independently of the image content and reflects the global properties of the image. To improve the accuracy of background light estimation, this paper utilizes the method introduced by [52] for estimating global background light. This method notes that there is a certain correlation between manually annotated global background light and the distribution of actual images in the

RGB channels. Based on this, linear and nonlinear transformations are applied to underwater images. Firstly, the average (Avg) and standard deviation (Std) of the global background light in the GB channels are linearly transformed. A linear regression model is then used to capture the correlation between the GB channels and the manually annotated background light:

$$B^{c'} = 1.13 \times Avg^{c'} + 1.11 \times Std^{c'} - 25.6 \tag{11}$$

here, $Avg^{c'}$ and $Std^{c'}$ are the average and standard deviation of the input image's $c'$ channel (referring to the GB channels), respectively. The coefficients and constant terms are determined through linear regression.

For the R channel, a nonlinear curve fitting model is employed:

$$B^r = 140 / (1 + 14.4 \times Exp(-0.034 \times Med_r)) \tag{12}$$

where, $Med_r$ represents the median of the R channel. The coefficients are obtained through nonlinear regression. To mitigate the impact of noise and extreme pixel values, the Avg, Std, and median are computed using the middle 80% of the intensity histogram of the coarse underwater images, excluding the lowest and highest 10% of values.

To prevent issues of overfitting or underfitting caused by extreme values, constraints are applied to the estimated global background light, restricting pixel values to the range of 5 to 250. The final estimated global background light is computed as follows:

$$B^c = \min(\max(B^{r,c'}, 5), 250), c \in \{r, g, b\} \tag{13}$$

Fig. 4 shows the different components and the reconstructed image during the image enhancement process of the Mamba-UIE framework. It is worth noting that our framework is data-driven, which allows for some tolerance of inaccurate estimations.

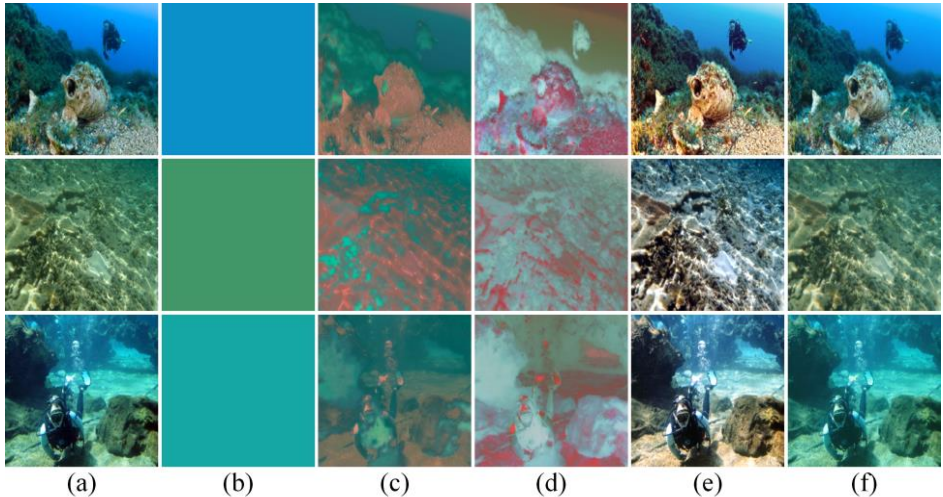

(a)    (b)    (c)    (d)    (e)    (f)

Fig. 4. Different components in the enhancement process. (a) Raw. (b) global background light. (c) direct transmission map. (d) backscattering transmission map. (e) scene radiance. (f) reconstructed image.

## 3.6 Loss function

**A Direct loss:** Our proposed Mamba-UIE primarily includes direct loss for scene radiance and reconstruction loss. Reconstruction loss emphasizes the consistency between the reconstructed image and the original image, thereby indirectly constraining the scene radiance. Direct loss mainly comprises $l_2$ loss, structural similarity (SSIM) loss $l_{SSIM}$, edge loss $l_{edge}$. $l_2$ loss compares the generated scene radiance $J$ with the reference $J_{label}$ at the pixel level, standardizing the details and textures of the generated image:

$$l_2 = \| J - J_{label} \|_2^2 \tag{14}$$

We add SSIM loss to regulate the similarity between the generated scene radiance and the reference.

$$l_{SSIM} = 1 - SSIM(J, J_{label}) \tag{15}$$

where $SSIM$ represents the structural similarity. $SSIM$ compares image patches based on luminance, contrast, and structure. The specific formula for $SSIM$ can be found in [53].

Edge loss is defined as:

$$l_{edge} = \sqrt{\| Lap(J) - Lap(J_{label}) \|^2 + \varepsilon^2} \tag{16}$$

where $Lap$ represents the Laplacian operator. $\varepsilon$ is an infinitesimal quantity. The edge loss enhances the fidelity and realism of high-frequency details. Additionally, we use the UIQM loss to improve the image's performance in terms of UIQM:

$$l_{UIQM} = 1/UIQM(J) \tag{17}$$

The specific formula for $UIQM$ can be found in [54].

**B Reconstruction Loss:** The input image is decomposed into scene radiance, global background light, direct transmission map, and backscattering transmission map. These four components are reassembled according to the formation model to obtain the reconstructed image. Applying appropriate constraints between the reconstructed image and the original image can improve the accuracy of scene radiance estimation. The reconstruction loss mainly consists of two parts: $l_2$ loss and SSIM loss. $l_2$ loss is defined as follows:

$$l^R_2 = \| I - I' \|_2^2 \tag{18}$$

where $I$ represents the original image and $I'$ represents the reconstructed image generated through formation model. The superscript $R$ denotes that it belongs to the reconstruction loss. SSIM loss is defined as follows:

$$l^R_{SSIM} = 1 - SSIM(I, I') \qquad (19)$$

Finally, our total loss function is:

$$l_{total} = l_2 + l_{SSIM} + \lambda l_{edge} + l_{UIQM} + l^R_2 + l^R_{SSIM} \qquad (20)$$

$\lambda$ is set to 0.05 according to [55].

## 4 Experiment and analysis

### 4.1 Setup

#### 4.1.1. Implementation details

The experiments were conducted on a desktop computer equipped with an Intel Core i9-10850K CPU, an NVIDIA GeForce RTX 3090 GPU with 24GB of VRAM, and 64GB of RAM. The project utilized the PyTorch framework and involved 200 training iterations. The batch size was set to 1. Our model was trained using the ADAM optimizer with a learning rate of $lr = 2 \times 10^{-8}$. We conducted comprehensive qualitative and quantitative evaluations on three specific representative public datasets.

#### 4.1.2. Datasets

We selected three representative datasets, UIEB [23], EUVP [56], and U45 [57], to train and test our model. These datasets are divided into two categories: 1) Full referenced datasets: 890 pairs of images from UIEB and 1,200 pairs of images from EUVP. 2) Non-referenced datasets: 60 challenging images (Challenging 60) from UIEB and 45 images from U45. To facilitate training and testing across different methods, the image sizes in the three datasets were uniformly scaled to 256×256. The dataset structure and divisions are shown in Table 1.

**Table 1.** The amount and distribution of the underwater datasets

| Datasets | Trian | | Test | |
|---|---|---|---|---|
| | Paired | Unpaired | Paired | Unpaired |
| UIEB | 800 | - | 90 | 60 |
| EUVP | 1000 | - | 200 | - |
| U45 | - | - | - | 45 |

#### 4.1.3. Compared methods

Our method was compared with the state-of-the-art methods. The traditional methods include: BTLM [52], UNTV [10], MLLE [7], HLRP [9], PCDE [58], WWPF [27], HFM [59]. Deep learning-based methods: Ucolor [13], PUIE [60], USUIR [47], U-shape [32], UDAformer [30], LiteEnhance [61], DICAM [62]. To ensure a fair comparison, all parameter settings in the comparison models follow the configurations provided in the original papers, except for necessary modifications related to image size.

**4.1.4. Evaluation metrics**

The full reference evaluation metrics and the non-reference evaluation metric were used to quantitatively evaluate and analyze the generated enhanced images. Full-reference evaluation metrics include MSE, PSNR [63], and SSIM [53]. MSE compares the differences between the generated image and the reference pixel by pixel. PSNR is a metric based on MSE that represents the difference between the generated image and the GT. SSIM compares images at the block level based on brightness, contrast, and structure. A smaller MSE value indicates better image quality, while higher PSNR and SSIM values indicate better image quality.

Since ground truths for underwater images are often unavailable, it is necessary to use no-reference evaluation metrics to quantify the imaging quality of underwater images. No-reference evaluation metrics mainly include UIQM [54] and UCIQE [64]. UIQM comprehensively evaluates images from color index, sharpness index, and contrast index. UCIQE provides a comprehensive evaluation of underwater images based on color concentration, saturation, and contrast. The higher the values of UIQM and UCIQE, the better the quality of the image.

**4.2 Quantitative evaluation**

Firstly, we conduct comparisons on full referenced datasets. The performance of different methods on the UIEB and EUVP datasets is shown in Table 3. It is worth noting that PUIE requires four types of labels, which are only provided for UIEB. Due to the lack of corresponding labels, we did not implement PUIE on EUVP. On the UIEB dataset, our proposed Mamba-UIE achieved the best results in MSE and PSNR metrics, and the second-best result in SSIM. On the EUVP dataset, Mamba-UIE achieved the second-best result in SSIM and the third-best results in MSE and PSNR. As for the UIQM metric, although Mamba-UIE did not rank in the top three, its performance was still quite good.

**Table 2.** the performance results of different methods on the UIEB and EUVP datasets. The numbers following the method indicate the year. The top three scores are highlighted in bold, with red for the best, blue for second place, and green for third place. This format is also applied to the tables in the following sections unless otherwise specified.

| Methods | UIEB | | | | EUVP | | | |
|---|---|---|---|---|---|---|---|---|
| | MSE($\times 10^3$)↓ | PSNR(dB)↑ | SSIM↑ | UIQM↑ | MSE($\times 10^3$)↓ | PSNR(dB)↑ | SSIM↑ | UIQM↑ |
| BTLM 2020 | 1.34±1.61 | 19.11±4.27 | 0.77±0.09 | 2.59±0.65 | 0.93±0.39 | 18.80±1.77 | 0.63±0.06 | 2.10±0.52 |
| UNTV 2021 | 1.59±1.00 | 16.80±2.42 | 0.56±0.11 | 1.94±0.71 | 1.13±0.65 | 18.20±2.25 | 0.54±0.08 | 1.83±0.73 |
| MLLE 2022 | 1.17±0.88 | 18.59±3.28 | 0.73±0.09 | 2.25±0.65 | 1.71±0.79 | 16.23±1.96 | 0.56±0.07 | 1.92±0.53 |
| HLRP 2022 | 2.97±1.06 | 13.68±1.54 | 0.23±0.08 | 2.71±0.79 | 3.58±1.62 | 12.95±1.74 | 0.14±0.04 | 2.81±0.70 |
| PCDE 2023 | 1.85±1.41 | 16.67±3.30 | 0.66±0.15 | 2.06±0.82 | 1.84±0.70 | 15.78±1.58 | 0.55±0.08 | 1.71±0.77 |
| WWPF 2023 | 1.05±0.79 | 18.95±3.04 | 0.78±0.08 | 2.52±0.51 | 1.41±0.66 | 17.05±1.89 | 0.60±0.06 | 2.37±0.39 |
| HFM 2024 | 0.65±0.52 | 21.22±3.41 | 0.82±0.07 | 2.83±0.42 | 0.71±0.07 | 20.50±2.24 | 0.66±0.38 | 3.04±0.26 |
| Ucolor 2022 | 0.30±0.34 | 25.08±3.79 | 0.91±0.06 | **3.20±0.33** | 0.24±0.17 | 25.16±2.68 | 0.79±0.06 | **3.30±0.27** |
| PUIE 2022 | 0.44±0.45 | 23.49±4.12 | 0.85±0.08 | **3.12±0.38** | - | - | - | - |
| USUIR 2022 | 0.34±0.38 | 25.09±4.59 | 0.86±0.06 | 3.07±0.34 | 0.50±0.35 | 21.94±2.62 | 0.68±0.05 | 2.67±0.21 |
| U-shape 2023 | 0.63±0.49 | 21.31±3.31 | 0.80±0.08 | **3.19±0.34** | 0.72±0.11 | 20.89±4.22 | **0.84±0.91** | **3.31±0.30** |
| UDAformer 2023 | **0.26±0.38** | **26.80±5.10** | **0.93±0.05** | 3.10±0.34 | **0.19±0.17** | **26.33±2.89** | **0.82±0.06** | **3.22±0.33** |
| LiteEnhance 2024 | 0.36±0.43 | 24.79±4.49 | 0.90±0.07 | 3.10±0.36 | 0.21±0.18 | 25.97±2.79 | **0.82±0.06** | 3.13±0.38 |
| DICAM 2024 | **0.28±0.37** | **27.01±5.84** | **0.93±0.06** | 3.09±0.39 | **0.19±0.17** | **26.42±2.88** | 0.82±0.07 | 3.13±0.40 |
| Mamba-UIE | **0.26±0.37** | **27.13±5.49** | **0.93±0.06** | 3.10±0.36 | **0.20±0.18** | **26.19±2.91** | **0.83±0.06** | 3.16±0.35 |

Due to the lack of paired images for training, we tested the models (only for supervised methods) trained on the UIEB dataset on no-reference datasets. The performance of different methods on the Challenging60 and U45 datasets is shown in Table 3. On the Challenging60 dataset, Mamba-UIE achieved the third-best result in the UIQM metric. On the U45 dataset, Mamba-UIE achieved the second-best result in the UIQM metric. The quantitative analysis across the three datasets demonstrates the effectiveness of our proposed Mamba-UIE.

Table 3. The performance results of different methods on Challenging60 and U45 datasets.

| Methods | Challenging60 | | U45 | |
|---|---|---|---|---|
| | UIQM↑ | UCIQE↑ | UIQM↑ | UCIQE↑ |
| BTLM 2020 | 1.91±0.53 | 0.59 | 2.52±0.75 | 0.60 |
| UNTV 2021 | 1.62±0.68 | 0.59 | 1.84±0.62 | **0.62** |
| MLLE 2022 | 1.96±0.55 | 0.59 | 2.60±0.50 | 0.59 |
| HLRP 2022 | 2.11±0.85 | **0.64** | 3.04±0.65 | **0.64** |
| PCDE 2023 | 1.74±0.71 | **0.60** | 2.18±0.99 | 0.61 |
| WWPF 2023 | 2.47±0.51 | 0.59 | 2.93±0.35 | 0.60 |
| HFM 2024 | 2.31±0.58 | 0.59 | 3.12±0.27 | 0.61 |
| Ucolor 2022 | **2.86±0.61** | 0.57 | **3.37±0.25** | 0.59 |
| PUIE 2022 | 2.74±0.54 | 0.56 | 3.28±0.28 | 0.57 |
| USUIR 2022 | 2.82±0.49 | **0.63** | 3.30±0.19 | **0.64** |
| U-shape 2023 | 2.85±0.56 | 0.54 | 3.30±0.26 | 0.56 |
| UDAformer 2023 | **2.90±0.53** | 0.59 | 3.24±0.30 | 0.61 |
| LiteEnhance 2024 | 2.69±0.61 | 0.56 | 3.26±0.36 | 0.58 |
| DICAM 2024 | 2.65±0.61 | 0.58 | **3.31±0.32** | 0.61 |
| Mamba-UIE | **2.72±0.54** | 0.59 | **3.32±0.24** | 0.61 |

### 4.3 Qualitative analysis in different scenarios

Due to the complexity of underwater environments, different degradation scenarios pose various challenges for underwater image enhancement. To validate the reliability and robustness of Mamba-UIE, we conducted qualitative analyses for four common underwater degradation scenarios. These scenarios include color offset scenes (Fig. 5 raw), blurry scene (Fig. 6 raw), scattering scenes (Fig. 7 raw), and extreme degradation scenes (Fig. 8 raw). The enhanced images generated by different methods are shown in Figs. 5-8.

The degree of light absorption in water depends on its wavelength. Red light attenuates the fastest, while blue light attenuates the slowest. This attenuation causes color offset in images, leading to green or blue phenomena, which are common degradation scenarios. As shown in Fig. 5, for underwater green scenes, most methods except BTLM, UNTV, HLRP, PCDE, HFM, USUIR, and LiteEnhance can effectively improve the underwater green scene. For underwater blue scenes, methods such as

Ucolor, PUIE, U-shape, UDAformer, LiteEnhance, DICAM, and Mamba-UIE can effectively restore the colors of underwater images. However, methods like MLLE, PCDE, and WWPF exhibit excessive enhancement.

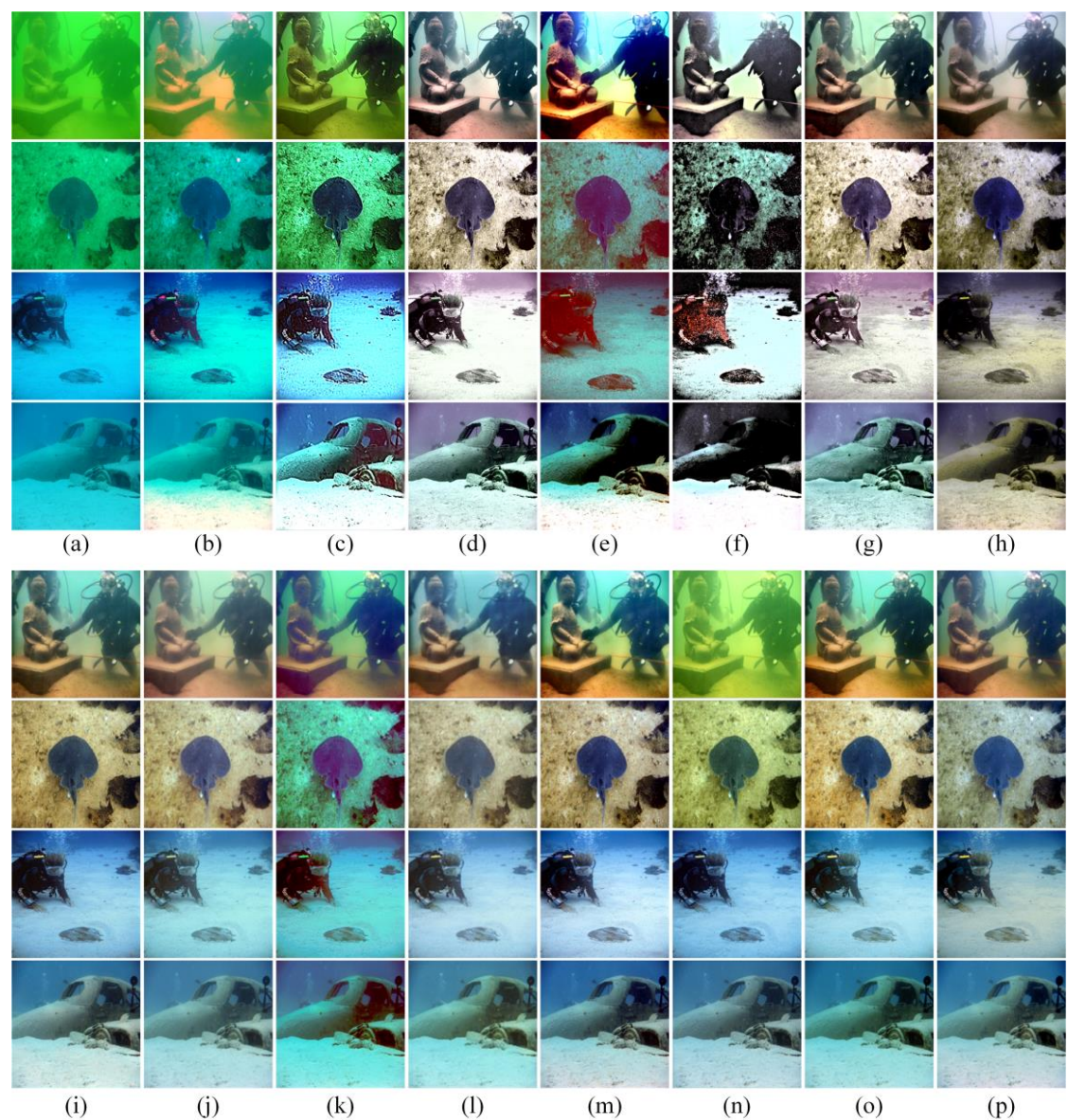

**Fig. 5.** Comparison of the color offset scenes from U45 dataset. (a) Raw. (b) BTLM. (c) UNTV. (d) MLLE. (e) HLRP. (f) PCDE. (g) WWPF. (h) HFM. (i) Ucolor. (j) PUIE. (k) USUIR. (l) U-shape. (m) UDAformer. (n) LiteEnhance. (o) DICAM. (p) Mamba-UIE.

Light is absorbed underwater, leading to reduced contrast and blurred images. Additionally, the presence of suspended particles in water exacerbates light scattering and absorption, increasing image blur. As shown in Fig. 6, methods such as UNTV, MLLE, WWPF, HFM, Ucolor, USUIR, UDAformer, LiteEnhance, DICAM, and Mamba-UIE effectively improve image blur and enhance image details. It is worth noting that due to the lack of effective local feature extraction, U-shape tends to exhibit blurriness in most scenes, as seen in other examples.

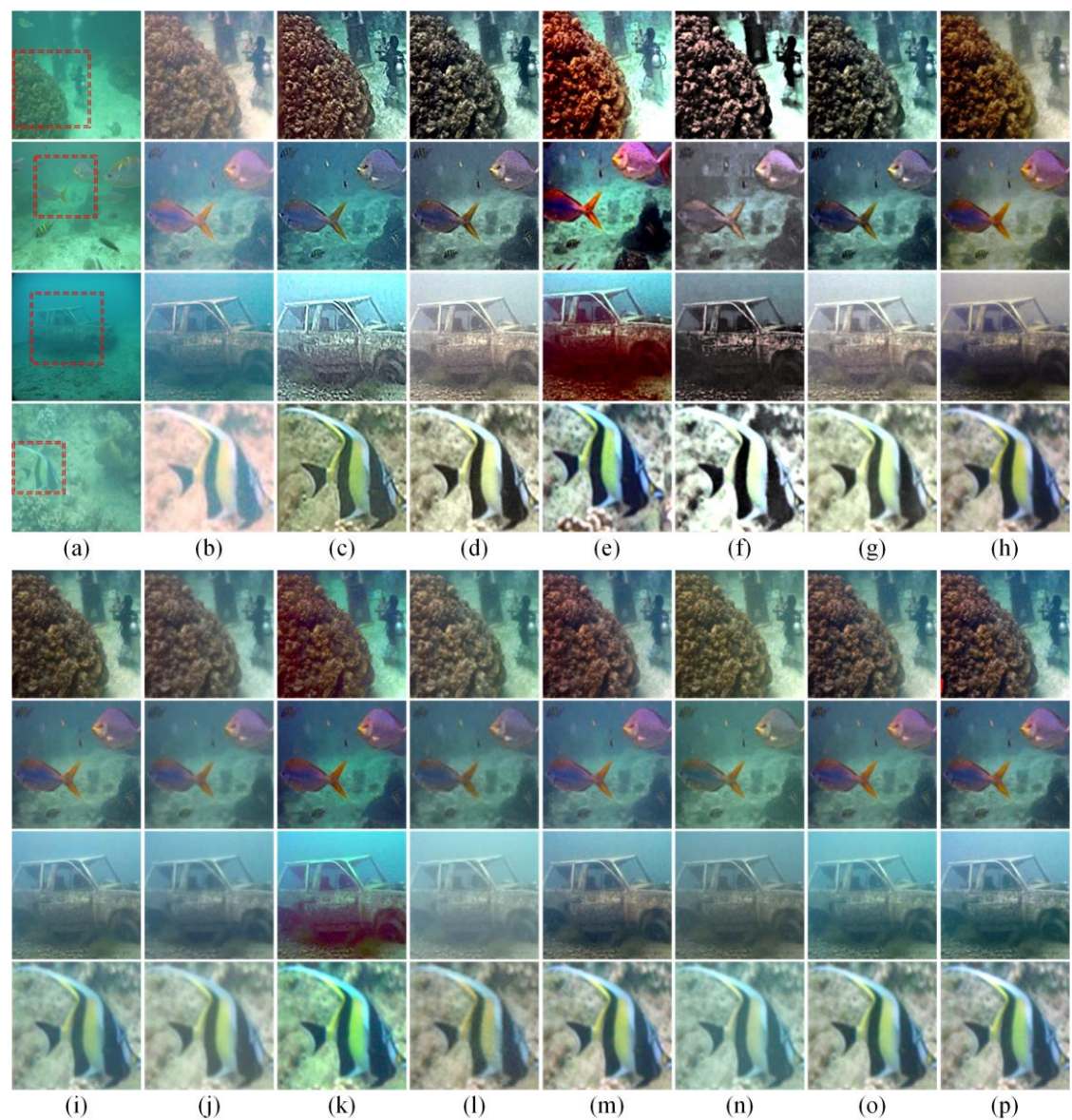

**Fig. 6.** Comparison of the blurry scene from UIEB dataset. (a) Raw. (b) BTLM. (c) UNTV. (d) MLLE. (e) HLRP. (f) PCDE. (g) WWPF. (h) HFM. (i) Ucolor. (j) PUIE. (k) USUIR. (l) U-shape. (m) UDAformer. (n) LiteEnhance. (o) DICAM. (p) Mamba-UIE.

When light passes through uneven media such as marine organisms, sediment particles, bubbles, or water masses, its propagation path becomes complex, altering the spatial distribution of light and leading to further scattering and Tyndall effects. This poses significant challenges for UIE. As shown in Fig. 7, in such scenarios, methods like BTLM, UNTV, HLRP, PCDE, and HFM cause severe distortion in underwater images. LiteEnhance leads to slight over-enhancement of the images, while other deep learning methods perform relatively well.

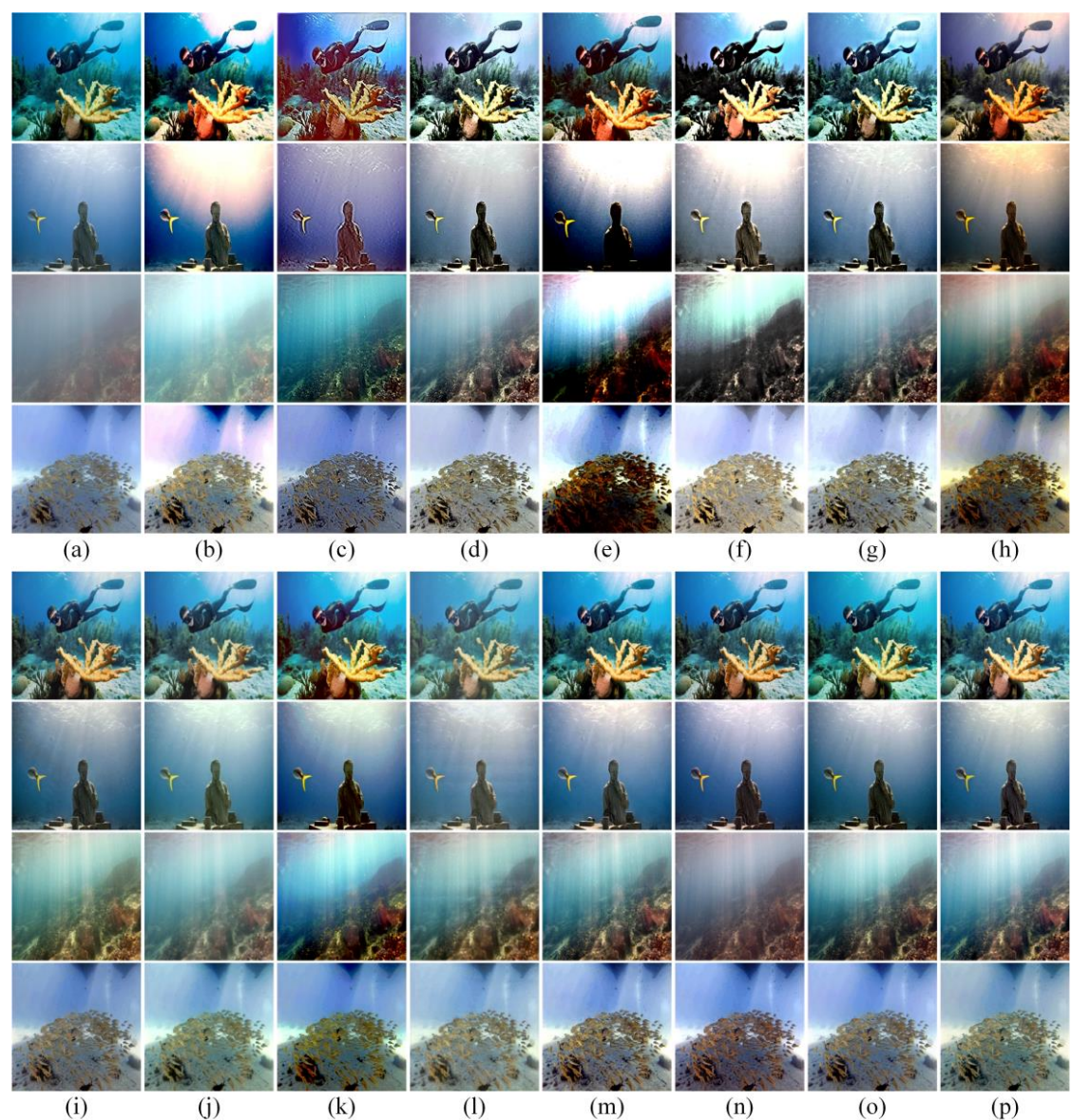

**Fig. 7.** Comparison of the scattering scenes from UIEB dataset. (a) Raw. (b) BTLM. (c) UNTV. (d) MLLE. (e) HLRP. (f) PCDE. (g) WWPF. (h) HFM. (i) Ucolor. (j) PUIE. (k) USUIR. (l) U-shape. (m) UDAformer. (n) LiteEnhance. (o) DICAM. (p) Mamba-UIE.

In extremely degraded environments with weak light and severe scattering from suspended particles, underwater images exhibit significantly low quality, including color distortion, low contrast, and blurriness. As shown in Fig. 8, traditional methods perform very poorly except for MLLE and WWPF. Ucolor exhibits over-enhancement. PUIE, USUIR, U-shape, UDAformer, LiteEnhance, and DICAM show inadequate color restoration. Relatively speaking, Mamba-UIE performs the best.

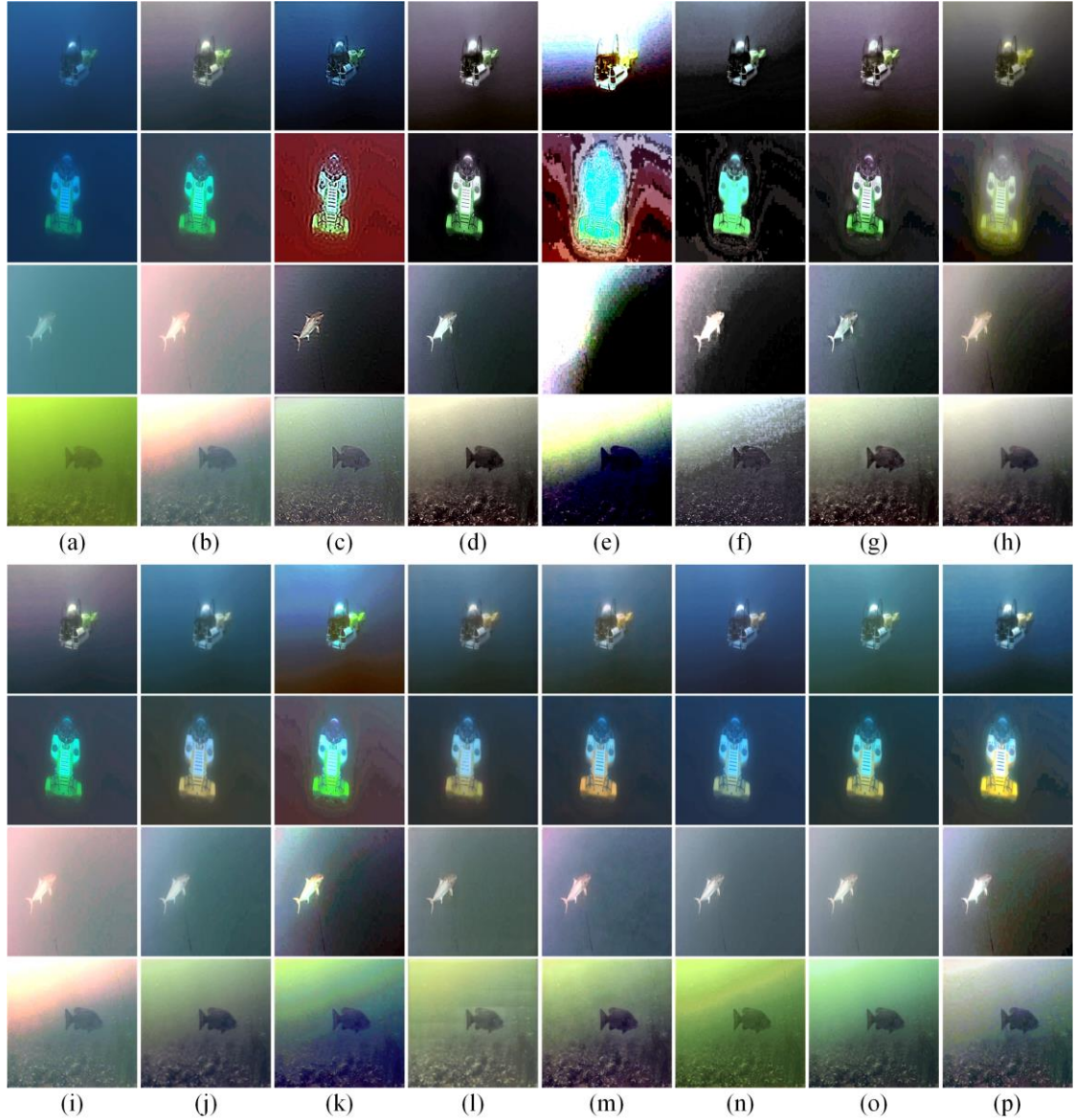

**Fig. 8.** Comparison of the extreme degeneration scenes from Challeng60. (a) Raw. (b) BTLM. (c) UNTV. (d) MLLE. (e) HLRP. (f) PCDE. (g) WWPF. (h) HFM. (i) Ucolor. (j) PUIE. (k) USUIR. (l) U-shape. (m) UDAformer. (n) LiteEnhance. (o) DICAM. (p) Mamba-UIE.

In the four common underwater degradation scenarios, our proposed Mamba-UIE achieved good enhancement results, demonstrating its effectiveness and robustness.

### 4.4 Histogram Comparison

Red light undergoes severe attenuation underwater, resulting in lower corresponding pixel values. Reflected in the histogram, this is seen as the red channel curve shifting to the left, as shown in Fig. 9(a). To more intuitively and comprehensively demonstrate the effectiveness of Mamba-UIE in restoring attenuated light, we calculated the cumulative RGB histograms for different enhancement methods on the UIEB test set, as shown in Fig. 9. Traditional methods and USUIR cause significant peaks in the histogram (frequency values on the vertical axis exceeding 100,000), indicating insufficient or excessive color enhancement. Our proposed Mamba-UIE has the smallest extremum on the vertical axis, not exceeding 40,000, with a more uniform RGB distribution. The generated images are closer to those taken in air. These demonstrate Mamba-UIE's exceptional capability in suppressing the dominant channel and compensating for the attenuated channels, effectively correcting color bias.

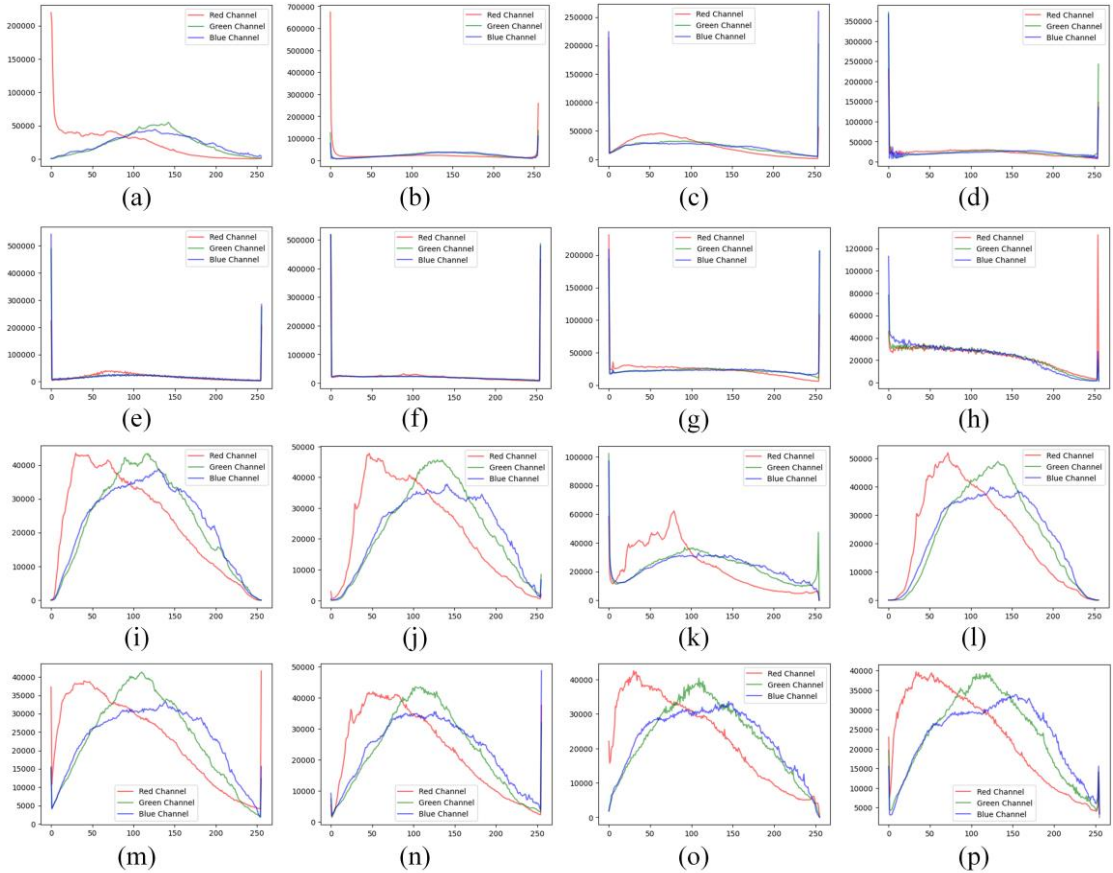

**Fig. 9.** Comparison of RGB histogram distributions on the UIEB dataset. The vertical axis represents frequency, and the horizontal axis represents pixel values. (a) Raw. (b) BTLM. (c) UNTV. (d) MLLE. (e) HLRP. (f) PCDE. (g) WWPF. (h) HFM. (i) Ucolor. (j) PUIE. (k) USUIR. (l) U-shape. (m) UDAformer. (n) LiteEnhance. (o) DICAM. (p) Mamba-UIE.

### 4.5 White balance test

To further demonstrate the performance of the proposed Mamba-UIE in color restoration, we conducted a white balance test. The test was performed on the Color-Checker7 [65] dataset, which consists of images taken by divers holding a standard color card in a swimming pool. The results are shown in Fig. 10. In this test, traditional methods exhibited poor color restoration and some overexposure issues. Most deep learning methods achieved better color restoration of underwater images. Mamba-UIE not only restored colors effectively but also enhanced the brightness of underwater images, improving underwater visibility.

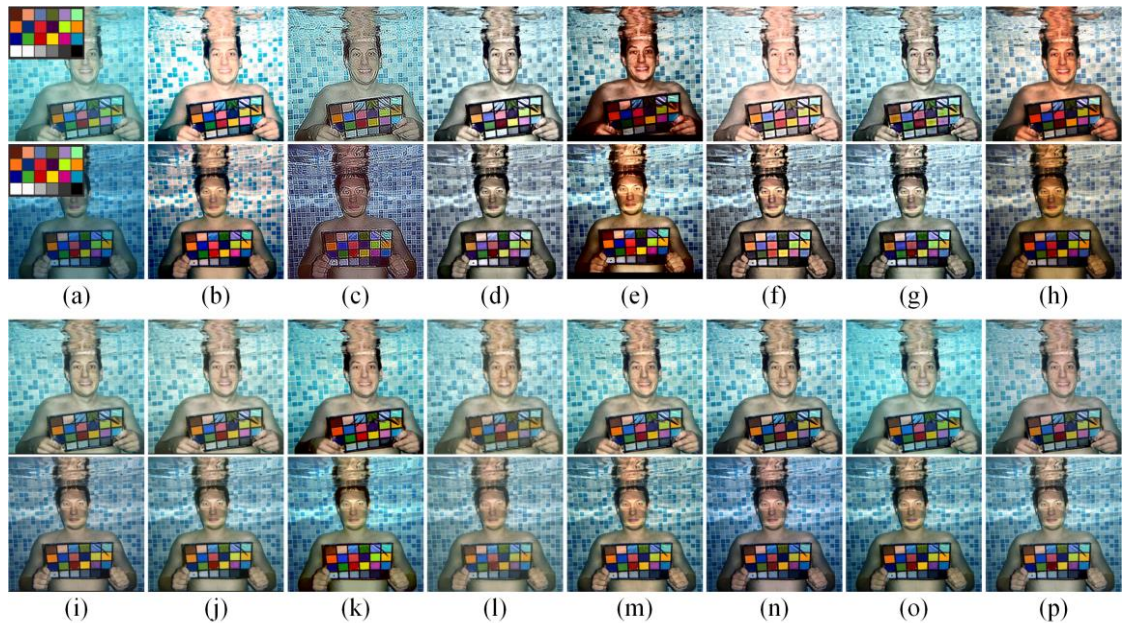

**Fig. 10.** White balance test results. (a) Raw. (b) BTLM. (c) UNTV. (d) MLLE. (e) HLRP. (f) PCDE. (g) WWPF. (h) HFM. (i) Ucolor. (j) PUIE. (k) USUIR. (l) U-shape. (m) UDAformer. (n) LiteEnhance. (o) DICAM. (p) Mamba-UIE. A standard color card is included in the upper-left of (a) Raw.

### 4.6 Ablation study

### 4.6.1 Effectiveness of physical model constraint

Our proposed Mamba-UIE framework consists of two main components: the introduced physical model constraint and the Mamba-UIE network. To demonstrate the effectiveness of the introduced physical model constraint, we conducted ablation experiments. As shown in Table 4, even without the Mamba component, the proposed framework still achieves good performance, and the inclusion of Mamba further enhances the performance.

Table 4. Comparison of results with and without different components on the UIEB. "w/o Mamba" indicates removing the Mamba parallel branch from the J-net network and using the CNN backbone network only as J-net. "w/o physical constraint" indicates removing the physical constraint, i.e., not using the reconstruction loss. The top two scores are highlighted in bold, with red for the best and blue for the second-best.

| Methods | MSE($\times 10^3$)↓ | PSNR(dB)↑ | SSIM↑ | UIQM↑ |
| --- | --- | --- | --- | --- |
| Mamba-UIE | **0.26±0.37** | **27.13±5.49** | **0.93±0.06** | 3.10±0.36 |
| w/o Mamba | 0.27±0.32 | 26.12±4.62 | **0.92±0.05** | **3.14±0.34** |
| w/o physical constraint | **0.27±0.31** | **26.15±4.68** | 0.92±0.05 | **3.12±0.34** |

**4.6.2 Constraint based on different formation models**

Our proposed Mamba-UIE is constrained by physical formation principles, skillfully combining deep learning methods with prior knowledge. We explored the effectiveness of Mamba-UIE based on different formation model constraints, specifically including the Koschmieder light scattering model [50]:

$$I(x) = J(x)T(x) + (1-T(x))A \tag{21}$$

In the Koschmieder light scattering model, direct scattering and backscattering images are not distinguished. A unified transmission map $T(x)$ is used. This represents a simplified underwater image formation model.

Retinex model [66]:

$$I(x) = R(x)L(x) \tag{22}$$

The Retinex model assumes that an underwater image can be represented as the product of illumination $L(x)$ and reflection $R(x)$. and, $R(x)$ is the clean underwater image that we aim to restore.

Jaffe-McGlamery [67, 68] model:

$$I(x) = J(x)T(x) + (1-T(x))A + (J(x)T(x))*g(x) \tag{23}$$

The Jaffe-McGlamery model suggests that an underwater image consists of three components: direct projection $J(x)T(x)$, forward scattering $(1-T(x))A$, and background scattering $(J(x)T(x))*g(x)$, where $*$ denotes convolution and $g(x)$ represents the point spread function. We used a trainable 9×9 convolutional kernel as a replacement for $g(x)$. Ablation experiments based on different physical model constraints are shown in Table 5.

**Table 5.** Ablation experiment results based on different physical model constraints on the UIEB dataset. The top three scores are highlighted in bold, with red for the best and blue for the second-best.

| Formation Models | MSE($\times 10^3$)↓ | PSNR(dB)↑ | SSIM↑ | UIQM↑ |
|---|---|---|---|---|
| Mamba-UIE | **0.26±0.37** | **27.13±5.49** | **0.93±0.06** | **3.10±0.36** |
| Koschmieder | 0.27±0.33 | 26.54±5.21 | 0.92±0.06 | 3.04±0.38 |
| Retinex | 0.31±0.42 | 26.50±5.53 | 0.92±0.06 | **3.09±0.41** |
| Jaffe-McGlamery | **0.26±0.31** | **26.85±5.12** | **0.93±0.05** | 3.06±0.37 |

It can be seen that the Jaffe-McGlamery model achieves performance closest to that of Mamba-UIE according to Table 5. This is because the Jaffe-McGlamery model accounts for background scattering, which is one of the primary causes of image blur. On the other hand, our proposed Mamba-UIE framework can be relatively easily extended to other formation models. Even when using other physical models for constraints, it still achieves good results. This demonstrates the effectiveness of using physical models to constrain the supervised learning process. Fig. 11 provides examples of enhanced images based on different imaging models, and their results are quite similar.

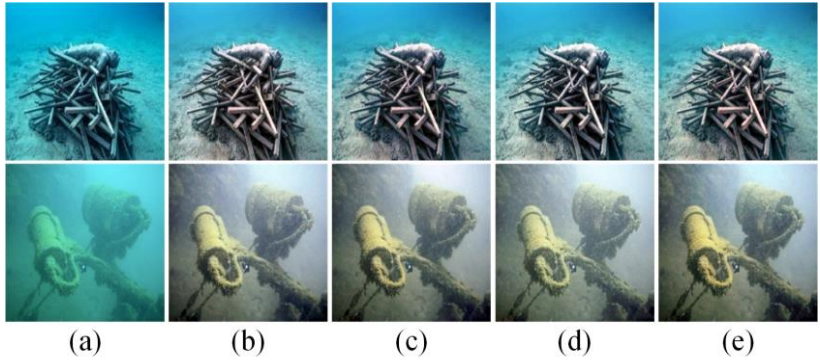

**Fig. 11.** Enhanced images based on different formation models. (a) Raw. (b) Mamba-UIE. (c) Koschmieder-based. (c) Retinex-based. (d) Jaffe-McGlamery-based.

## 5 Conclusion

We propose a physical model constraint-based underwater image enhancement framework, Mamba-UIE. Specifically, we decompose the input image into four components: underwater scene radiance, direct transmission map, backscatter transmission map, and global background light. These components are reassembled according to the revised underwater image formation model, and a reconstruction consistency constraint is applied between the reconstructed image and the original image, thereby achieving effective physical constraint on the underwater image enhancement process. To address the quadratic computational complexity of Transformers when handling long sequences, we introduce the Mamba-UIE network based on linear complexity SSM. By incorporating the Mamba in Convolution block, long-range dependencies are modeled at both the channel and spatial levels, while the CNN backbone is retained to recover local features and details. Extensive experiments demonstrate that our proposed Mamba-UIE achieves state-of-the-art performance, effectively enhancing underwater images across various scenarios. Additionally, our framework can be easily extended to other underwater formation models, providing new idea for integrating deep learning with prior-based approaches.

## CRediT authorship contribution statement

**Song Zhang:** Writing-original draft, Visualization, Validation, Software, Methodology, Conceptualization. **Yuqing Duan:** Visualization, Validation, Software. **Daoliang Li:** Writing–review & editing, Funding acquisition. **Ran Zhao:** Writing–review & editing, Supervision, Methodology, Funding acquisition, Conceptualization.

## Declaration of competing interest

The authors declare that they have no known competing financial interests or personal relationships that could have appeared to influence the work reported in this paper.

## Data availability

Data will be made available on request.

## Acknowledgments

This paper was supported by The National Natural Science Foundation of China (NO.32273188) and Key Research and Development Plan of the Ministry of Science and Technology (NO.2022YFD2001700).